\documentclass[
]{ceurart}

\usepackage{listings}
\usepackage{todonotes}

\begin{document}

\conference{}

\title[mode= title]{SFMGNet: A Physics-based Neural Network To Predict Pedestrian Trajectories}

\author[1]{Sakif Hossain}[%
orcid=0000-0003-0800-8392,
email=sakif.hossain@tu-clausthal.de,]
\address[1]{Department of Informatics, Clausthal University of Technology, Julius-Albert-Str. 4, 38678 Clausthal-Zellerfeld, Germany}
\author[1]{Fatema T. Johora}[%
email=fatema.tuj.johora@tu-clausthal.de,
]
\author[1]{J\"org P. M\"uller}[%
email=joerg.mueller@tu-clausthal.de,
]
\author[1]{Sven Hartmann}[%
orcid=0000-0003-4565-9645,
email=sven.hartmann@tu-clausthal.de,
]
\author[1]{Andreas Reinhardt}[%
email=andreas.reinhardt@tu-clausthal.de,
]
\begin{abstract}
  Autonomous robots and vehicles are expected to soon become an integral part of our environment. Unsatisfactory issues regarding interaction with existing road users, performance in mixed-traffic areas and lack of interpretable behavior remain key obstacles. To address these, we present a physics-based neural network, based on a hybrid approach combining a social force model extended by group force (SFMG) with Multi-Layer Perceptron (MLP) to predict pedestrian trajectories considering its interaction with static obstacles, other pedestrians and pedestrian groups. We quantitatively and qualitatively evaluate the model with respect to realistic prediction, prediction performance and prediction "interpretability". Initial results suggest, the model even when solely trained on a synthetic dataset, can predict realistic and interpretable trajectories with better than state-of-the-art accuracy.
\end{abstract}



\maketitle

\section{Introduction}
There has been growing research interest in autonomous technologies like autonomous vehicles, service robots, goods carriers, and surveillance robots. It is generally expected that such autonomous entities will be a part of our daily environment in the near future. However, prior to this, several hard issues remain to be resolved. Firstly, their performance in mixed-traffic zones and interaction with existing road users (e.g. cars, pedestrians, cyclists) requires further inspection \cite{talebpour2016influence}, \cite{bhavsar2017risk}. Secondly, uncertainty or lack of interpretability in their motion behavior and decisions make them less "socially acceptable" \cite{bhavsar2017risk}. Again, in a mixed-traffic zone autonomous robots/vehicles need to predict future trajectories of other road users (cars, pedestrians, cyclists, etc.) to plan their own trajectories and navigate safely. However, doing so is a complex task due to overlapping interactions among road users, road user groups and static obstacles.

\begin{figure}
  \centering
  \includegraphics[width = 0.60\textwidth]{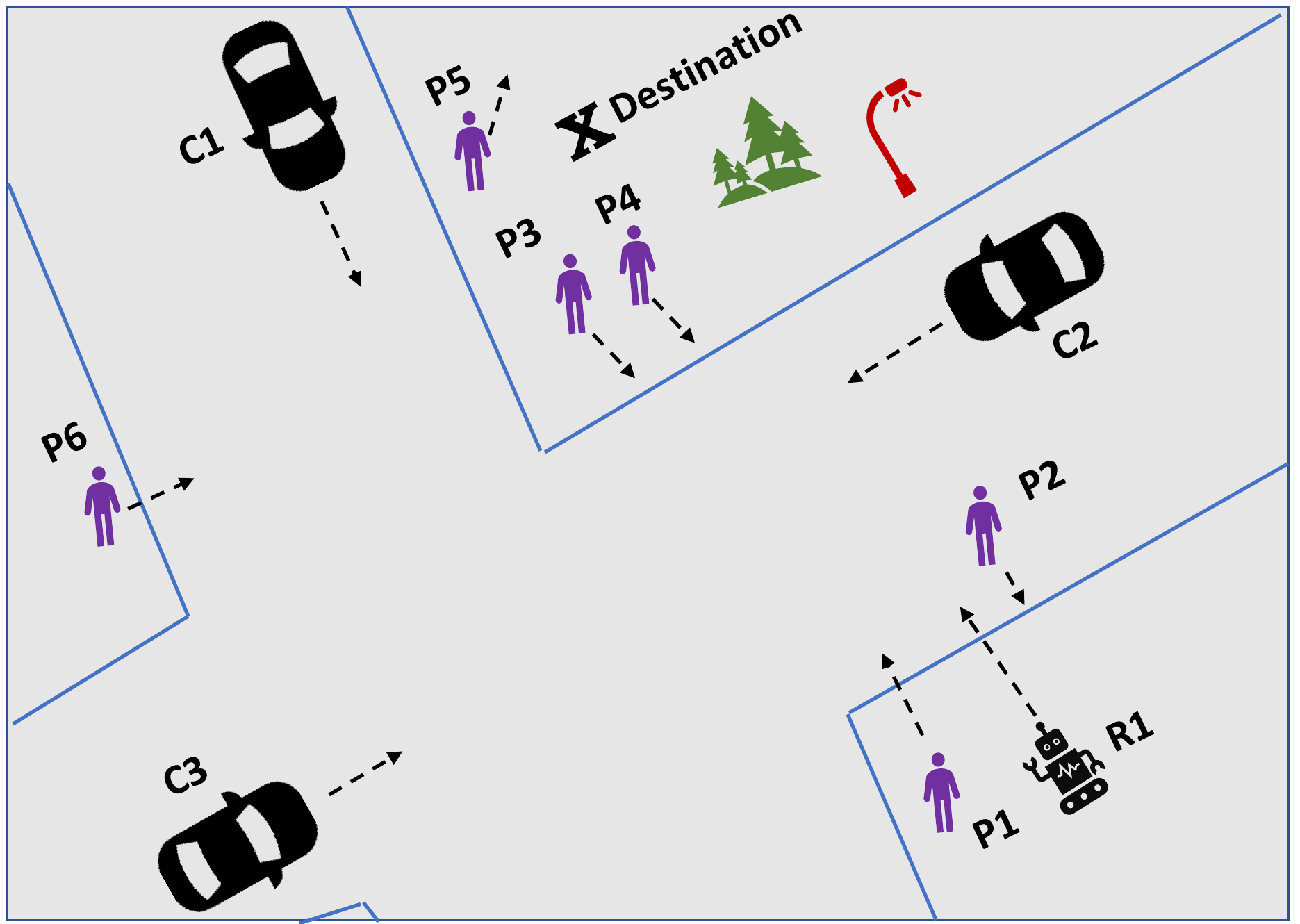}
  \caption{A robot placed in a mixed-traffic zone traveling towards a destination.}
  \label{fig: example_scenario}
\end{figure}

Figure \ref{fig: example_scenario} shows an example scenario. Service robot R1 is placed in a mixed-traffic zone, trying to navigate through other road users to reach its destination X. On its path, it will encounter cars (C2), pedestrians (P1, P2, P3 and P4) and static obstacles (e.g. trees, lamps). For that, it may need to predict the trajectories of these other road users and plan its own path based on that. However, the behavior of an individual road user can also depend on other factors present in the environment. Factors like interaction with static obstacles, other road users, and (if the road user is traveling in a group) other group members. So, the robot/vehicle has to consider all these factors while predicting other road users' trajectories and planning its own path. 


As pedestrians are vulnerable road users and form the majority in most mixed-traffic zones, modeling pedestrian behavior warrants high importance. This task entails estimating a pedestrian's future motion within seconds based on its previous motion and environment related information. 

The main contribution of this paper is a novel pedestrian motion prediction model, a prerequisite for autonomous robot/vehicle path planning, which considers pedestrian interaction with static obstacles, other pedestrians and pedestrian groups. We take a hybrid approach combining the classical social force model (SFM) \cite{helbing1995social} extended by pedestrian group behavior modeling \cite{ahmed2019investigating}, with a neural network for predicting pedestrian motion.

The reminder of this paper is arranged as follows: Section \ref{sec:related_work} reviews related work on trajectory prediction. Section \ref{sec:sfmg} then briefly discusses SFMG; Our model architecture is described in Section \ref{sec:physics_based_model}, followed by its evaluation in Section \ref{sec:evaluation}. Section \ref{sec:conclusion} concludes and discusses future research opportunities.

\section{Related work}
\label{sec:related_work}
Broadly\nocite{hossainthesis}, road user trajectory prediction algorithms can be divided into theory-based or physics-based approaches and learning-based approaches. The first group of approaches use domain or theoretical knowledge and/or physics-based models to model road user motion. The social force model \cite{helbing1995social} and its various extensions (\cite{yang2018social}, \cite{pascucci2018should}, \cite{rinke2017multi}, \cite{anvari2015modelling}, \cite{ahmed2019investigating}, \cite{johora2020zone}) are widely used for this purpose. At its core, SFM defines pedestrian motion as being motivated by external forces exerted upon it by the static obstacles and other road users. These methods can model road user motion realistically and they are interpretable. However, complexity of estimating model parameters hinders their scalability and they require too much explicit expert involvement, which in turn restricts their suitability to include new road user types \cite{cheng2020trajectory}. 

Learning based approaches using machine learning algorithms, especially artificial neural networks (ANN) have the potential to overcome above-mentioned issues. With sufficient amount of relevant data, a machine learning algorithm can be trained to estimate underlying motion dynamics of a road user. Different types of neural network architectures have been used to predict pedestrian trajectory, including Multi-layer Perceptron \cite{scholler2020constant}, Long Short-Term Memory (LSTM) and Gated Recurrent Unit (GRU) (\cite{wang2017capturing}, \cite{alahi2016social}, \cite{cheng2018modeling}), Conditional Variational Auto-encoders \cite{cheng2020amenet}, Generative Adversarial Networks \cite{gupta2018social}, Graph Neural Networks \cite{vemula2018social}, and Graph Attention Networks \cite{huang2019stgat} and so on. However, the size of neural network needed to estimate complex human behavior and the number of training samples needed jointly become a major bottleneck. Moreover, neural networks tend to be like "black-boxes", lacking explainability. Explainability is required for autonomous entities to analyze their behavior, attribute responsibility and to engender trust \cite{pasquale2017toward}. Lack of a prior model in an ANN makes explainability hard or nearly impossible. There have some attempts to combine theory-based and learning-based approaches to avail the advantages of both and consequently design a "Grey-box" model \cite{kroll2000grey}. 

In \cite{johora2020agent}, GSFM, an agent-based model, and LSTM-DBSCAN, a learning-based model,  were combined in a hierarchical manner, i.e. GSFM takes over when LSTM-DBSCAN predicts conflicting trajectories. In \cite{kim2020modeling}, the authors combine an ANN with a cellular automata model to estimate traffic congestion. Here an ANN helps to estimate the cellular automata model's parameters. In \cite{karpatne2017theory}, Karpatne et. al. propose a new paradigm of approaches which aim to combine theory-based knowledge with data-driven approaches and avail subsequent advantages. In \cite{antonucci2020generating}, authors propose a combination between SFM and MLP to predict pedestrian trajectories. Here, the neural network architecture has been designed in such a manner that it estimates SFM parameters. This work \cite{antonucci2020generating} can estimate the acceleration force towards the desired destination and repulsive forces exerted by the nearby obstacles using two different MLPs. Interestingly, the second MLP consists both acceleration force part and repulsive force part together in a single network, jeopardizing its interpretable nature. Moreover, the authors do not consider interaction with other road users. Another approach to combine SFM and neural networks to design a differentiable simulation model is introduced in \cite{kreiss2021deep}. Here, the neural network is used to estimate the interaction potentials. Pedestrian interaction with obstacle/s and other pedestrian/s are considered by the authors. The authors did not test the model \cite{kreiss2021deep} on real data. \textbf{Noticeably}, none of these works consider interactions among pedestrians and pedestrian group/s, although pedestrian groups contribute toward 70\% of the road users \cite{moussaid2010walking}.

\paragraph{}Therefore, in this work we combine SFMG and neural network to model pedestrian motion and, consider interactions with obstacles, other pedestrians and pedestrian groups. Here SFMG helps understand and model/predict a pedestrian's motivation behind making a certain motion behavior (i.e. through corresponding forces). The neural network architecture mimics the SFMG equations and it is solely used to estimate the equation parameters. SFMG serves as a prior model for the neural network. Thus, the overall approach retains social force model's interpretability. Again, we employ individual networks to estimate individual forces and combine them in a \textit{modular} manner to get total force. Refer to Section \ref{sec:physics_based_model} for further details on the model architecture.
\section{Social force model with pedestrian groups (SFMG)}
\label{sec:sfmg}
In this section, we briefly discuss the social force model \cite{helbing1995social} to understand how we can incorporate it with Multi-Layer Perceptron (MLP). Suppose, a certain pedestrian $\alpha$ is moving towards its destination $\overrightarrow{x}_{a}^o$ at a certain desired velocity $\overrightarrow{w}_\alpha(t) := v_\alpha^o \overrightarrow{e}_\alpha(t)$ at time t. Its motion path can be disturbed by static obstacles in the environment (e.g. walls, trees, etc.) and other road users. Then, according to SFM \cite{helbing1995social}, the total force acting upon $\alpha$ is defined as: 

\begin{equation}
    \frac{d\overrightarrow{v_{\alpha}}(t)}{dt} = \overrightarrow{f}_\alpha^o + \sum_\beta \overrightarrow{f}_{\alpha\beta}^{SOC} + \sum_B \overrightarrow{f}_{\alpha B}^{SOC} + \sum_i \overrightarrow{f}_{\alpha i}^{SOC} + \xi
    \label{eq:overall_force}
\end{equation}

Again, the acceleration force towards the destination, $\overrightarrow{f}_\alpha^o$ is defined as: 
\begin{equation}
    \overrightarrow{f}_{a}^o  =  \frac{v_\alpha^o \overrightarrow{e}_\alpha(t) - \overrightarrow{v}_\alpha(t)}{\tau}
    \label{eq:acceleration_force}
\end{equation}

Here, $\overrightarrow{v}_\alpha(t)$ is the current velocity, $\tau$ is relaxation time and $\overrightarrow{e}_\alpha(t) = \frac{\overrightarrow{x}_{a}^n - \overrightarrow{x}_{a}}{||\overrightarrow{x}_{a}^n - \overrightarrow{x}_{a}||}$ is the desired direction. $\overrightarrow{x}_{a}^n$ and $\overrightarrow{x}_{a}$ denote next position and current position of pedestrian $\alpha$ respectively. Pedestrian's tendency to keep a safe distance from static obstacles is denoted by the repulsive force $\overrightarrow{f}_{\alpha B}^{SOC}$ as: 
\begin{equation}
    \overrightarrow{f}_{\alpha B}^{SOC}  = U_{\alpha B}^o e^{-{d}_{\alpha B} (t)/{R}}  \overrightarrow{\eta}_{\alpha B}
    \label{eq:sfm boundary_force}
\end{equation}

Here, B denotes a static obstacle and $U_{\alpha B}^o$ is the interaction strength between $\alpha$ and obstacle B. ${d}_{\alpha B}$ and $\overrightarrow{\eta}_{\alpha B}$ are the distance and the direction unit vector between $\alpha$ and B respectively. The repulsive force between $\alpha$ and another pedestrian $\beta$ is denoted by:
\begin{equation}
    \overrightarrow{f}_{\alpha\beta}^{SOC} = V_{\alpha\beta}^o e^{-{d}_{\alpha\beta} (t)/{\sigma}}  \overrightarrow{\eta}_{\alpha\beta} F_{\alpha\beta}
    \label{eq:sfm ped_force}
\end{equation}
where, ${d}_{\alpha\beta}$ and $\overrightarrow{\eta}_{\alpha\beta}$ are distance and unit direction vector between $\alpha$ and $\beta$. $F_{\alpha\beta}$ represents the anisotropic behavior of a pedestrian and its defined as:
\begin{equation}
    F_{\alpha\beta} = \lambda_\alpha + (1 - \lambda_\alpha)(1 + \cos{(\phi_{\alpha\beta})})/{2}
    \label{eq:F_alpha_beta}
\end{equation}
Here, $\lambda_a$ is a constant and $\phi_{\alpha\beta}$ is the angle between the motion direction of pedestrian $\alpha$ and the vector pointing in the direction of $\alpha$ to $\beta$. i.e. $\cos{(\phi_{\alpha\beta})} = \overrightarrow{\eta}_{\beta\alpha}(t). \overrightarrow{e}_\alpha(t)$. Combining equation \ref{eq:sfm ped_force} and \ref{eq:F_alpha_beta}, and simplifying it we get:
\begin{equation}
    \overrightarrow{f}_{\alpha\beta}^{SOC} = V_{\alpha\beta}^o (A_1 e^{{-{d}_{\alpha\beta} (t)}/{\sigma}}  \overrightarrow{\eta}_{\alpha\beta} + A_2 \overrightarrow{\eta}_{\alpha\beta} \overrightarrow{\eta}_{\alpha\beta} \overrightarrow{e}_\alpha (t) e^{{-{d}_{\alpha\beta} (t)}/{\sigma}} )
    \label{eq:ped_force_final}
\end{equation}
where, $A_1 = \lambda_\alpha + ({1 - \lambda_\alpha})/{2}$ and $A_2 = ({1 - \lambda_\alpha})/{2}$ are constants. 

The attractive force towards points of interests (e.g. friends, street artist, etc.) is given by $\overrightarrow{f}_{\alpha i}^{SOC}$. We ignore this force in our work to avoid further complexity. The classical SFM has been extended by \cite{ahmed2019investigating}, \cite{kremyzas2016towards} to include pedestrian group interaction by introducing the group force $\overrightarrow{f}_{group}$ as:
\begin{equation}
    \overrightarrow{f}_{group} = \overrightarrow{f}_{vis} + \overrightarrow{f}_{att}
    \label{eq: group_force}
\end{equation}

In above equation, $\overrightarrow{f}_{vis}$ denotes the pedestrian's desire to keep other group members within his/her field of view (FOV).  $\overrightarrow{f}_{att}$ is an attraction force representing the pedestrian's motivation to maintain group coherence. The terms $\overrightarrow{f}_{vis}$ and $\overrightarrow{f}_{att}$ can be described as:
\begin{equation}
    \overrightarrow{f}_{vis} = S_{vis} * \theta * \overrightarrow{V}_{desired}
    \label{eq: f_vis}
\end{equation}
\begin{equation}
    \overrightarrow{f}_{att}=\begin{cases} S_{att} *{ \overrightarrow{n}( A_{ij},C_i) } , & \text{if $dist( A_{ij},C_i)\geq d \: and \: \overrightarrow{V}_{desired} \neq 0$}\\
0, & \text{otherwise}.
\end{cases}
\label{eq: f_att}
\end{equation}

Here $\theta$ is the minimum rotation angle. $S_{vis}$ and $S_{att}$ are global strength parameters. $\overrightarrow{V}_{desired}$ is the desired velocity of any pedestrian $A_{ij}$ in group $G_i$. $\overrightarrow{n}( A_{ij},C_i)$ represents the normalized unit length vector between $A_{ij}$ and group centroid $C_i$. Adding group force and omitting $\overrightarrow{f}_{\alpha i}^{SOC}$ equation \ref{eq:overall_force} becomes: 

\begin{equation}
        \frac{d\overrightarrow{v_{\alpha}}(t)}{dt} = \overrightarrow{f}_\alpha^o + \sum_\beta \overrightarrow{f}_{\alpha\beta}^{SOC} + \sum_B \overrightarrow{f}_{\alpha B}^{SOC} +  \overrightarrow{f}_{group} + \xi
    \label{eq: final_force}
\end{equation}

We use the force equations of social force model with pedestrian group behavior modeling described in this section to design our pedestrian motion prediction model architecture.

\section{Physics-based neural network (SFMGNet)}
\label{sec:physics_based_model}

\begin{figure}
  \centering
  \includegraphics[width = 0.85\textwidth]{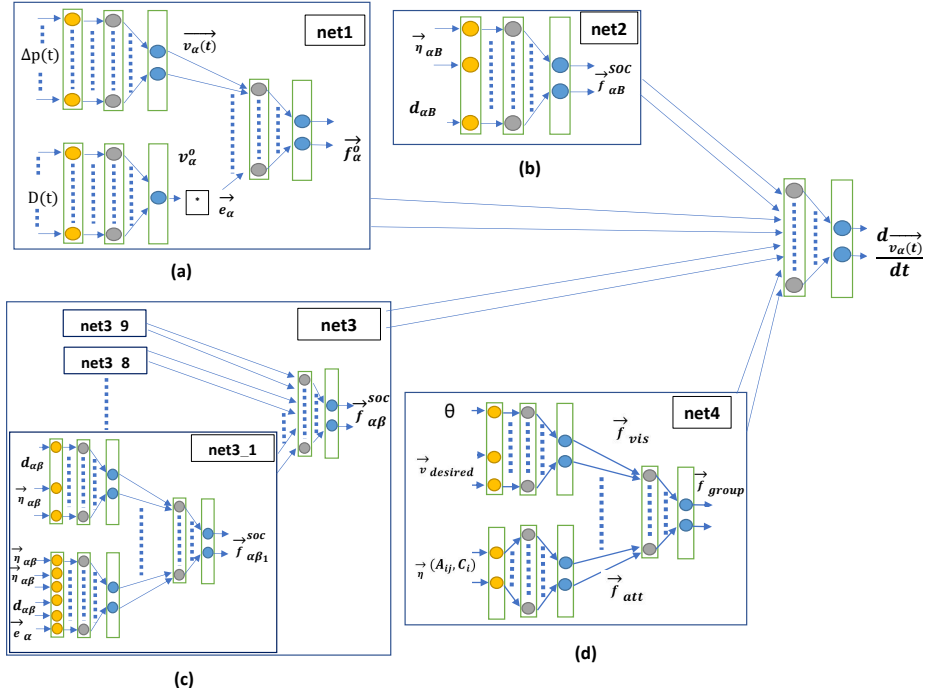}
  \caption{SFMGNet architecture. Here, net1 (a), net2 (b), net3 (c) and net4 (d) are the individual networks corresponding to the forces in Equation \ref{eq: final_force}. Note, net3 contains nine instances like net3\_1.}
  \label{fig:sfmnet}
\end{figure}

We aim to design our neural network based framework in such a way that it estimates the total force acting upon a pedestrian according to Equation \ref{eq: final_force}. Equation \ref{eq: final_force} aggregates four individual forces, namely acceleration force ($\overrightarrow{f}_\alpha^o$), repulsion from boundaries ($\overrightarrow{f}_{\alpha B}^{SOC}$), repulsive force from other pedestrians ($\overrightarrow{f}_{\alpha\beta}^{SOC}$) and group force ($\overrightarrow{f}_{group}$). We employ four individual networks to estimate these four forces separately. We call these individual networks as \textit{modules}. A final network combines these individual modules to give final force (${d\overrightarrow{v_{\alpha}}(t)}/{dt}$). Figure \ref{fig:sfmnet} shows the overall model architecture and highlights the individual modules. The model specific hyperparameters (e.g. activation function, learning rate, etc.) were chosen empirically. Further details are presented in Section \ref{subsec:individual_modules} and Section \ref{subsec:final_module}.

\subsection{Problem statement}
We formulate our trajectory prediction problem similar to standard approaches used in the literature \cite{vemula2018social}, \cite{sun2020reciprocal}. Given trajectories of all pedestrians in a scene, we need to predict their future trajectories. Let $\textbf{X} = [X_1, X_2, ..., X_N]$ be the current trajectories of all pedestrians present in the scene. We need to predict future trajectories, $\hat{Y} = [\hat{Y_1}, \hat{Y_2}, ..., \hat{Y_N}]$. Here, for a certain pedestrian $\alpha$, the given trajectories are $X_\alpha = [{x_\alpha}^t, {y_\alpha}^t]$ at time-steps $t = 0, 1, 2, ..., T$. The real or ground truth future trajectories for $\alpha$ are $Y_\alpha = [{x_\alpha}^t, {y_\alpha}^t]$ at time-steps $t = T + 1, ..., T_e$.

\subsection{Individual modules}
\label{subsec:individual_modules}

\paragraph{net1: Acceleration}This module estimates the goal directed acceleration force ($\overrightarrow{f}_\alpha^o$) as defined by Equation \ref{eq:acceleration_force}. The model architecture is designed according to Equation \ref{eq:acceleration_force}. This module is largely based on \cite{antonucci2020generating}. The module inputs are n previous positions (p) of the pedestrian $\alpha$, till current time-step t. The position values are normalized by subtracting the first position value of the window ($p^{(t-n)}$) to avoid spatial bias. The new position values are termed as $\Delta p(t)$. 

net1 consists of two parts: one sub-network to estimate the instantaneous velocity $\overrightarrow{v}_\alpha(t)$ and second sub-network estimates the desired velocity $v_\alpha^0 \overrightarrow{e}_\alpha(t)$. The inputs to the networks are $\Delta p(t)$ and $D(t)$ respectively. Here $D(t)$ refers to the instantaneous velocity magnitudes and it is calculated as: $D(t) = [ ||{\Delta p^{'}(t)}_1||,...., ||{\Delta p^{'}(t)}_{n-1}|| ]$. Where, $\Delta p^{'}(t) = \Delta p(t) - \Delta p(t - 1)$. The first sub-network consists of two MLPs, where the first MLP is given \textit{sigmoid} activation and second MLP rescales the outputs. The second sub-network also has two MLPs in it, with \textit{tanh} activation for the first MLP and second MLP rescales the outputs. This output is multiplied with goal directed unit vector, $\overrightarrow{e}_\alpha(t)$, which is estimated by the approach described in Section \ref{subsec:goal_prediction}. Finally, the two sub-networks outputs are aggregated by another MLP to give $\overrightarrow{f}_\alpha^o$. Figure \ref{fig:sfmnet} (a) shows the net1 architecture. The net1 architecture can be expressed as:
\begin{equation}
    \overrightarrow{f}_\alpha^o = sig(D(t) W_{vd})W_{vds} \overrightarrow{e}_\alpha(t) - tanh(\Delta p(t) W_{vi}) \cdot W_{vis}
    \label{eq:net1 final}
\end{equation}
Here $W_{vd}$, $W_{vds}$, $W_{vi}$ and $W_{vis}$ are weight matrices. This net1 architecture mimics Equation \ref{eq:acceleration_force}. 
\paragraph{net2: Repulsive force from static obstacles} This network is responsible for estimating the repulsive force from obstacles \cite{antonucci2020generating}. In this work, we only consider the nearest obstacle point. This force is defined by Equation \ref{eq:sfm boundary_force}. So, net2 architecture is designed according to this equation. Figure \ref{fig:sfmnet} (b) shows net2 architecture. The distance between pedestrian $\alpha$ and obstacle B ($d_{\alpha B}$), and direction vector pointing towards B to $\alpha$ ($\overrightarrow{\eta}_{\alpha B}$) are network inputs. net2 consists of two MLPs, where the first one is followed by a \textit{sigmoid} activation and another MLP rescales the outputs. net2 architecture can be expressed as:
\begin{equation}
    \overrightarrow{f}_{\alpha B}^{SOC} = sig(W_U e^{d_{\alpha B}/W_R} \overrightarrow{\eta}_{\alpha B}) \cdot W_{fB}
    \label{eq:net2}
\end{equation}
Here, $W_U$, $W_R$ and $W_{fB}$ are weight matrices. This structure mimics SFMG force Equation \ref{eq:sfm boundary_force}.
\paragraph{net3: Repulsive force from other pedestrians} This section focuses on estimating the repulsive forces experienced by a pedestrian from other pedestrians. Equation \ref{eq:ped_force_final} defines this force. So, this corresponding network aims to mimic this equation. The input parameters are $\overrightarrow{\eta}_{\alpha \beta}$, $d_{\alpha \beta}$ and $\overrightarrow{e}_\alpha$. $A_1$, $A_2$, $\sigma$ and $V_{\alpha \beta}^o$ are the learn-able parameters. Figure \ref{fig:sfmnet} (c) shows the network architecture used to estimate repulsive force exerted by one neighboring pedestrian. This network consists of two sub-networks. First sub-network takes $\overrightarrow{\eta}_{\alpha \beta}$ and $d_{\alpha \beta}$ as inputs and it has one MLP which is followed by \textit{relu} activation. This sub-section estimates first part of Equation \ref{eq:ped_force_final}. Second sub-network estimates second part of Equation \ref{eq:ped_force_final} with inputs: $\overrightarrow{\eta}_{\alpha \beta}$, $d_{\alpha \beta}$ and $\overrightarrow{e}_\alpha$. It also has similar architecture as first sub-network. Finally, outputs of these two sub-networks are combined by another MLP, to get $\overrightarrow{f}_{\alpha\beta}$ force. The architecture is expressed as:
\begin{multline}
    \overrightarrow{f}_{\alpha\beta}^{SOC} = (relu(W_{A1} \exp{[d_{\alpha\beta}(t)/W_{\sigma}]}\overrightarrow{\eta}_{\alpha\beta}) + 
    \\
    relu(W_{A2} \overrightarrow{\eta}_{\alpha\beta} \overrightarrow{\eta}_{\alpha\beta} \overrightarrow{e}_\alpha (t) \exp{[{-{d}_{\alpha\beta} (t)}/{W_{\sigma 2}}]})) \cdot W_{\alpha \beta}
\end{multline}
Here, $W_{A1}$, $W_{A2}$, $W_{\sigma}$, $W_{\sigma 2}$ and $W_{\alpha \beta}$ are weight matrices. This structure mimics Equation \ref{eq:ped_force_final}.

Again, one pedestrian can have more than one nearby pedestrian exerting repulsive force on it. So, we empirically consider a maximum nine nearby pedestrians for a certain pedestrian. We employ instances of the network described above for finding repulsive forces from them. These networks are named as net3\_i, where i = 1, 2, 3, ..., 9. A final network, net3 is used to recombine the outputs of these networks to get total repulsive force from other pedestrians. 

\paragraph{net4: Group force} net4 approximates the force exerted by pedestrian groups. The group force is given by Equation \ref{eq: group_force}. Group force has two parts: visibility force ($f_{vis}$) and attraction force ($f_{att}$). So, net4 also has two sub-networks to estimate these two forces. Figure \ref{fig:sfmnet} (d) shows the net4 architecture. The first network inputs are $\theta$ and $\overrightarrow{v}_{desired}$. It has two MLPs: first MLP is followed by \textit{relu} activation and second MLP rescales the \textit{relu} output. The second sub-network has similar architecture, with $\overrightarrow{n}( A_{ij},C_i)$ as input. The two outputs are combined by another MLP to estimate the final group force. This can be expressed as: 
\begin{equation}
    \overrightarrow{f}_{group} = ( \textit{relu}(W_{g1} \theta \overrightarrow{V}_{desired}) + \textit{relu}(W_{g2} \overrightarrow{n}( A_{ij},C_i))) \cdot W_G
    \label{eq: net4}
\end{equation}
Here, $W_{g1}$, $W_{g2}$ and $W_G$ are weight matrices. This structure is similar to Equation \ref{eq: group_force}.
\subsection{Module: Final recombination}
\label{subsec:final_module}
This final recombination module aggregates the outputs of net1, net2, net3 and net4 to predict total force  $\frac{d\overrightarrow{v_{\alpha}}(t)}{dt}$. This network consists of two MLPs. First MLP is followed by \textit{relu} activation and second MLP rescales this output. Figure \ref{fig:sfmnet} shows the overall architecture. Expressed as:
\begin{equation}
    \frac{d\overrightarrow{v_{\alpha}}(t)}{dt} = \textit{relu}( (\overrightarrow{f}_\alpha^o + \sum_\beta \overrightarrow{f}_{\alpha\beta}^{SOC} +  \overrightarrow{f}_{\alpha B}^{SOC} +  \overrightarrow{f}_{group} ) W_{IF}) \cdot W_{FF}
    \label{eq:net final}
\end{equation}
here, $W_{IF}$ and $W_{FF}$ are weight matrices. Note, as we consider only nearest obstacle point, we do not need to sum repulsive forces from static obstacles. If we compare, the overall architecture described by Equation \ref{eq:net final} is similar to Equation \ref{eq: final_force}.

\subsection{Goal prediction}
\label{subsec:goal_prediction}
Estimated goal or destination is required as goal directed unit vector ($\overrightarrow{e}_\alpha$) is an input for net1 and net3. In an environment with other pedestrians and static obstacles, a pedestrian's previous positions cannot be exploited to estimate $\overrightarrow{e}_\alpha$. So, we use Multiple Model Approach (MMA) \cite{bar2004estimation} to estimate it. Specifically, we use an Interaction Multiple Model (IMM) estimator. Different goal hypotheses based Kalman filters are chosen as filters for the IMM estimator. The goal hypotheses were generated using Constant turn rate and velocity (CTRV) \cite{blackman1999design} model. Then, the IMM estimator compares the different filter generated trajectories with the observations and chooses the most likely one. 
\section{Evaluation}
\label{sec:evaluation}
In this section, we perform a set of experiments to qualitatively and quantitatively evaluate SFMGNet in terms of realistic (Section \ref{subsec:realistic_trajectory}), interpretable (Section \ref{subsec:interpretable_trajectory}) and accurate (Section \ref{subsec:quantitative_comparison} and Section \ref{subsec:ablation_study}) trajectory prediction. We discuss the experimental setup, dataset details and feature extraction process in Sections \ref{subsec:experimental_setup}, \ref{subsec:dataset_overview} and \ref{subsec:feature_extraction} respectively.

\subsection{Experimental setup}
\label{subsec:experimental_setup}
We train our model using a synthetic dataset. Only the models trained with synthetic dataset were used during the evaluation process, even while testing on real world datasets. We create the synthetic dataset using the trajectories simulated by a SFMG (SFM extended with group force) based simulator \footnote{SFM: \url{https://github.com/svenkreiss/socialforce/tree/579543a3abe22716835acfa0b7d2d57fc1c199b6}. Extended for groups.} in a closed environment (crossing passageways). We run 1000 simulations of 30 second duration sampled at 0.1 seconds. The simulation run scenarios were chosen randomly following a discrete uniform distribution. The following conditions were chosen to simulate scenarios rich with different types of interactions. The pedestrians' staring positions were chosen randomly among four possible starting zones with 7 to 10 meters distance to destination point. The number of pedestrians in a certain scenario were also chosen randomly between 2 to 10. The presence of a group of size 2 to 4 is chosen randomly. Relaxation time $\tau$ = 0.5, is constant. We divide the dataset with sample randomization into three sets: 50\% for training set, 25\% for development set and 25\% for testing set.     

We train the models net1, net2, net4, net3 and individual instances of net3 for different neighboring pedestrians (net3\_1, net3\_2, ..., net3\_9) with an Adam optimizer with a learning rate of 0.001 and batch size 16. Number of epochs were as chosen based on model performance on development set. The final recombination layer was also trained by Adam with 0.1 learning rate. The window of motion observation is empirically chosen as n = 10. We train and optimize model based on corresponding mean squared errors (MSE). The final MSE values for net1, net2, net3, net4 and SFMGNet are 0.0066, 0.5295, 0.1594, 0.0181 and 0.2640 units respectively. 

\subsection{Dataset overview}
\label{subsec:dataset_overview}
To assess SFMGNet's performance quantitatively, we test its performance on two benchmark datasets namely \textit{ETH }\cite{pellegrini2009you} dataset and \textit{UCY} \cite{lerner2007crowds} dataset. The ETH dataset consists of two scenes: \textit{ETH} and \textit{Hotel}. ETH scene consists of 360 pedestrians and 61 pedestrian groups. Hotel scene has 389 pedestrians and 41 groups. The UCY dataset has five scenes: namely \textit{zara01, zara02, students01}, \textit{students03} and \textit{uni\_examples}. zara01 has 148 scenes and 45 groups. Zara02 consists of 204 pedestrians and 58 groups. students03 scene consists of 434 pedestrians and 104 groups. students01 consists of 414 pedestrians and uni\_examples has 118 pedestrians. students01, students03 and uni\_examples scenes together are called UNIV. All the datasets are sampled at 0.4 seconds. All these scenes consist of multiple pedestrians being present in the same frame. We use the group and destination related information given by the work in \cite{amirian2020opentraj}. 


    

\subsection{Feature extraction}
\label{subsec:feature_extraction}
\begin{figure}
  \centering
  \includegraphics[width = 0.85\textwidth]{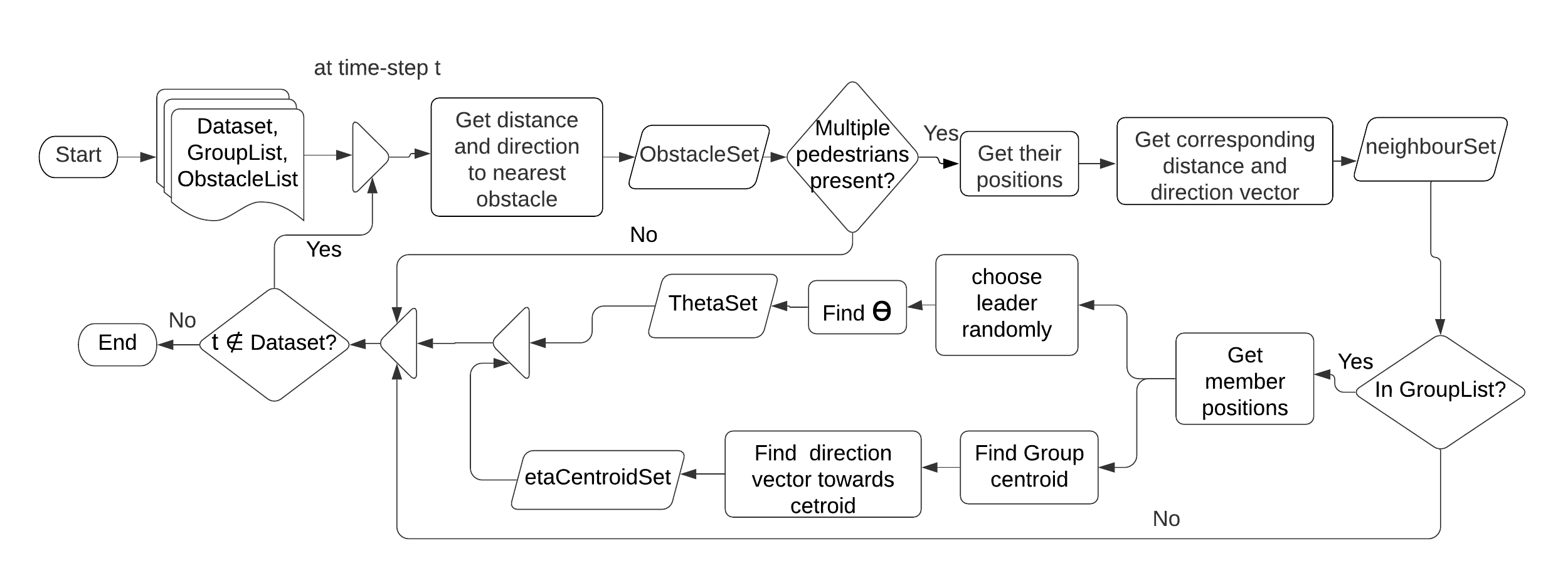}
  \caption{Flowchart for extracting features from a Dataset, given list of groups and obstacles.}
  \label{fig:feature_extraction}
\end{figure}

For a certain pedestrian $\alpha$ at time-step t, the SFMGNet model requires: distance and direction unit vector from nearest obstacle ($d_{\alpha B}$ and $\overrightarrow{\eta}_{\alpha B}$), distance and direction unit vector towards nearest (up to nine) pedestrian $\beta$ ($d_{\alpha \beta}$ and $\overrightarrow{\eta}_{\alpha \beta}$). If in a group, then it needs group Centroid directed unit vector ($\overrightarrow{\eta}(A_{ij}, C_i)$) and minimum rotation angle ($\theta$). However, a trajectory dataset normally consists of time-step, velocity and positions. Additionally, information about group formation \cite{amirian2020opentraj} and obstacles are provided (or extracted). So, we follow the approach depicted in the flowchart in Figure \ref{fig:feature_extraction} to extract features. A dataset (D), group (GroupList) and obstacle (ObstacleList) information are given to the algorithm. Then, at every time-step t in dataset we follow the following steps. First, nearest obstacle is taken from ObstacleList, distance from pedestrian and $\overrightarrow{\eta}_{\alpha B}$ is calculated and stored in an array (ObstacleSet). Next, we check if for multiple pedestrians' presence. If true, we treat as neighbors to ego pedestrian ($\alpha$) and, calculate and store $d_{\alpha \beta}$ and $\overrightarrow{\eta}_{\alpha \beta}$ in NeighbourList. Next, we check existence of groups from GroupList. If true, first we get the corresponding group member positions. Now, we need $\overrightarrow{\eta}(\alpha, C_i)$ and $\theta$. For calculating $\overrightarrow{\eta}(A_{ij}, C_i)$, first, we find the group Centroid based on group member positions and find their corresponding $\overrightarrow{\eta}(A_{ij}, C_i)$. Next, we store it in an array (etaCentroidList). For $\theta$, first we randomly choose a group leader, then we calculate $\theta$ based on group leader and other member positions (see \cite{ahmed2019investigating}, \cite{kremyzas2016towards} for further details), and store it in an array (ThetaSet).      

\subsection{Realistic trajectory prediction}
\label{subsec:realistic_trajectory}
\begin{figure}
  \centering
  \includegraphics[width = 0.55\textwidth]{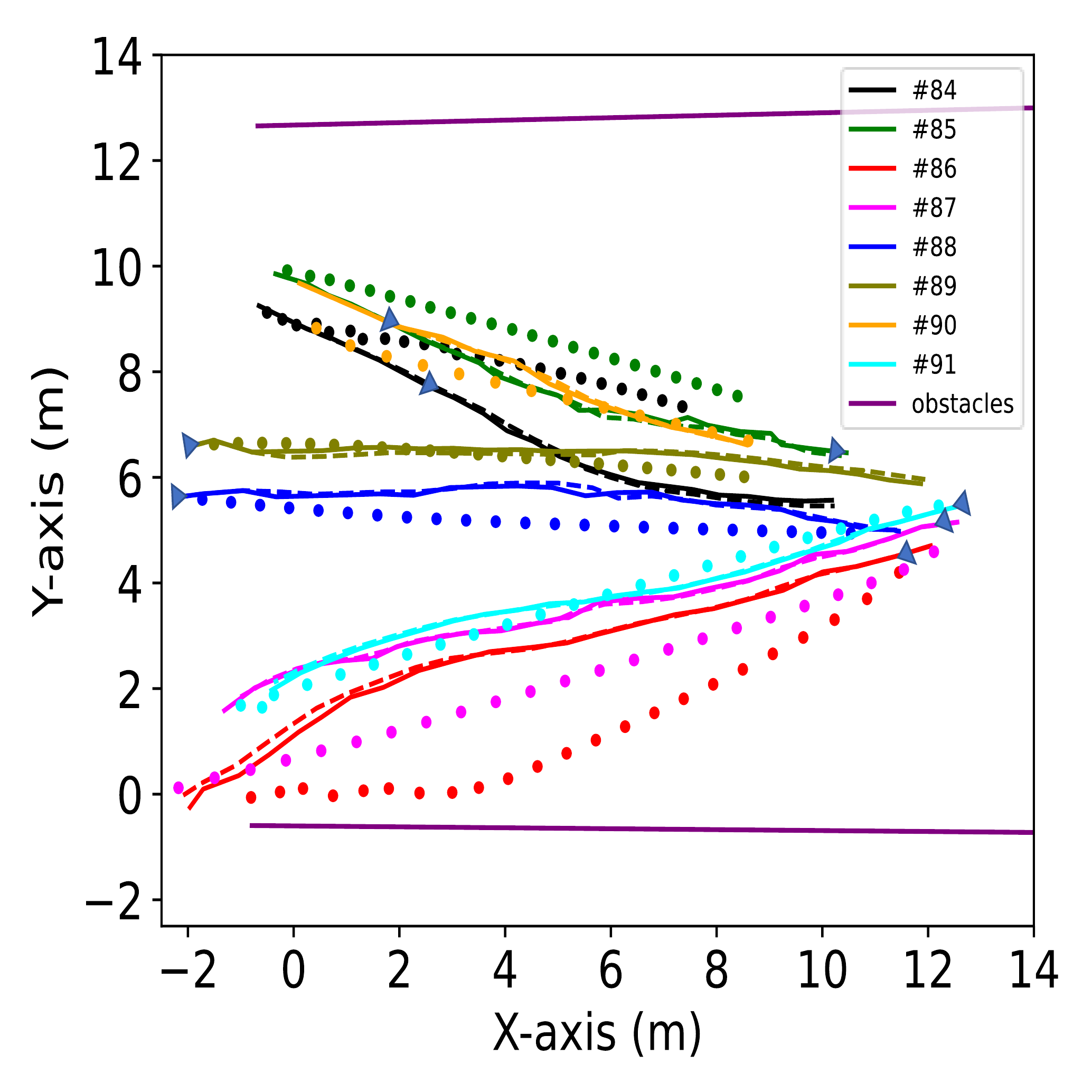}
  \caption{Real (solid lines), predicted (dashed lines) and SFMG simulated (dotted lines) trajectories of eight pedestrians. Here, arrows indicate the motion directions of the pedestrians.}
  \label{fig:all_ped_trajectories}
\end{figure}
\begin{figure}
  \centering
  \includegraphics[width = 0.90\textwidth]{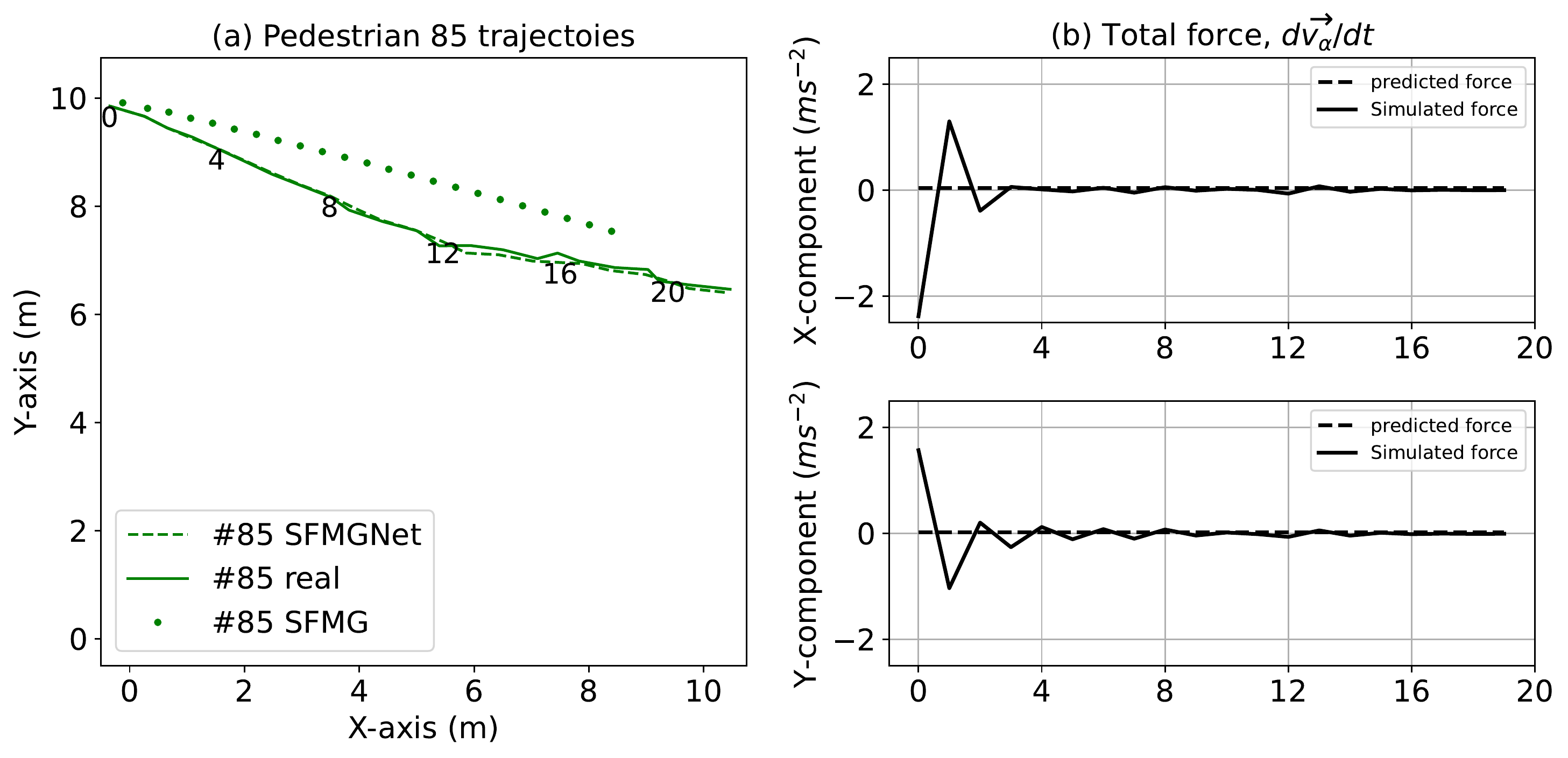}
  \caption{(a) Real, predicted and SFMG simulated trajectories of a pedestrian. Here, the numbers indicate the corresponding time-steps. (b) Total predicted force acting on the pedestrian.}
  \label{fig:ped6_trajectories}
\end{figure}

The aim of this section is to evaluate SFMGNet's ability to predict realistic trajectories through a case study. For this we choose a sample scenario (time-step: 4630 - 4880) from ETH scene from ETH dataset \cite{pellegrini2009you} with multiple number of pedestrians and pedestrian groups. This scenario consists of eight pedestrians namely: 84, 85, 86, 87, 88, 89, 90 and 91. Here, 86, 87 and 91 are traveling from right to left direction. And other pedestrians are traveling towards the opposite direction. 84 and 85 are in a pedestrian group. Similarly, 88 and 89 are also in a group. We predict all the pedestrians the trajectories with SFMGNet by taking 1.2 seconds of previous motion information as input and predicting next 4.8 seconds of trajectory (if possible) at a time. We also simulate their trajectories with SFMG \cite{helbing1995social} \cite{ahmed2019investigating} for comparison. Figure \ref{fig:all_ped_trajectories} shows the real (solid lines), predicted (dashed lines) and SFMG simulated (dots) trajectories. The arrows show the corresponding pedestrian's motion direction. We see that, predicted trajectories are similar to the real pedestrian trajectories, barring a couple of time-steps. The SFMGNet is able to predict and model slight deviations in every pedestrian's motion accurately. On the contrary, SFMG simulated trajectories are flatter in nature, not covering the influences of environment related influences on the pedestrian. Also, SFMG simulated pedestrian trajectories fail to reach or they overtake the destination point. As an example, we take a look at 85 in Figure \ref{fig:ped6_trajectories} (a). Here the numbers represent the time-steps. Our model predicted trajectory matches with the real trajectory and reaches the destination. Whereas, SFMG simulated trajectory is flatter and fails to reach the destination.

\subsection{Interpreting trajectory prediction}
\label{subsec:interpretable_trajectory}
In this section, we aim to interpret the SFMGNet predicted trajectories to a certain degree as a proof of concept. For that, we look at the predicted and simulated forces acting upon a certain pedestrian and try to establish a causal relationship with the environment elements/influences. We choose the pedestrian, 85 for this evaluation. Figure \ref{fig:all_ped_trajectories} shows the overall scenario, Figure \ref{fig:ped6_trajectories} zooms into 85 trajectories (a) and, predicted and SFMG simulated total force (b). Looking into the predicted total force (Figure \ref{fig:ped6_trajectories} (b)), we see an almost constant force acting upon the pedestrian, approximately 0.01 units in X-axis and 0.01 units in Y-axis. It points to the idea, that the pedestrian maintains a rather stable trajectory with subtle direction changes in its journey. The SFMG simulated force also remains stable apart from a spike at the beginning. However, the total force does not paint the whole picture. So, we explore the individual forces. 

\begin{figure}
  \centering
  \includegraphics[width = 0.90\textwidth]{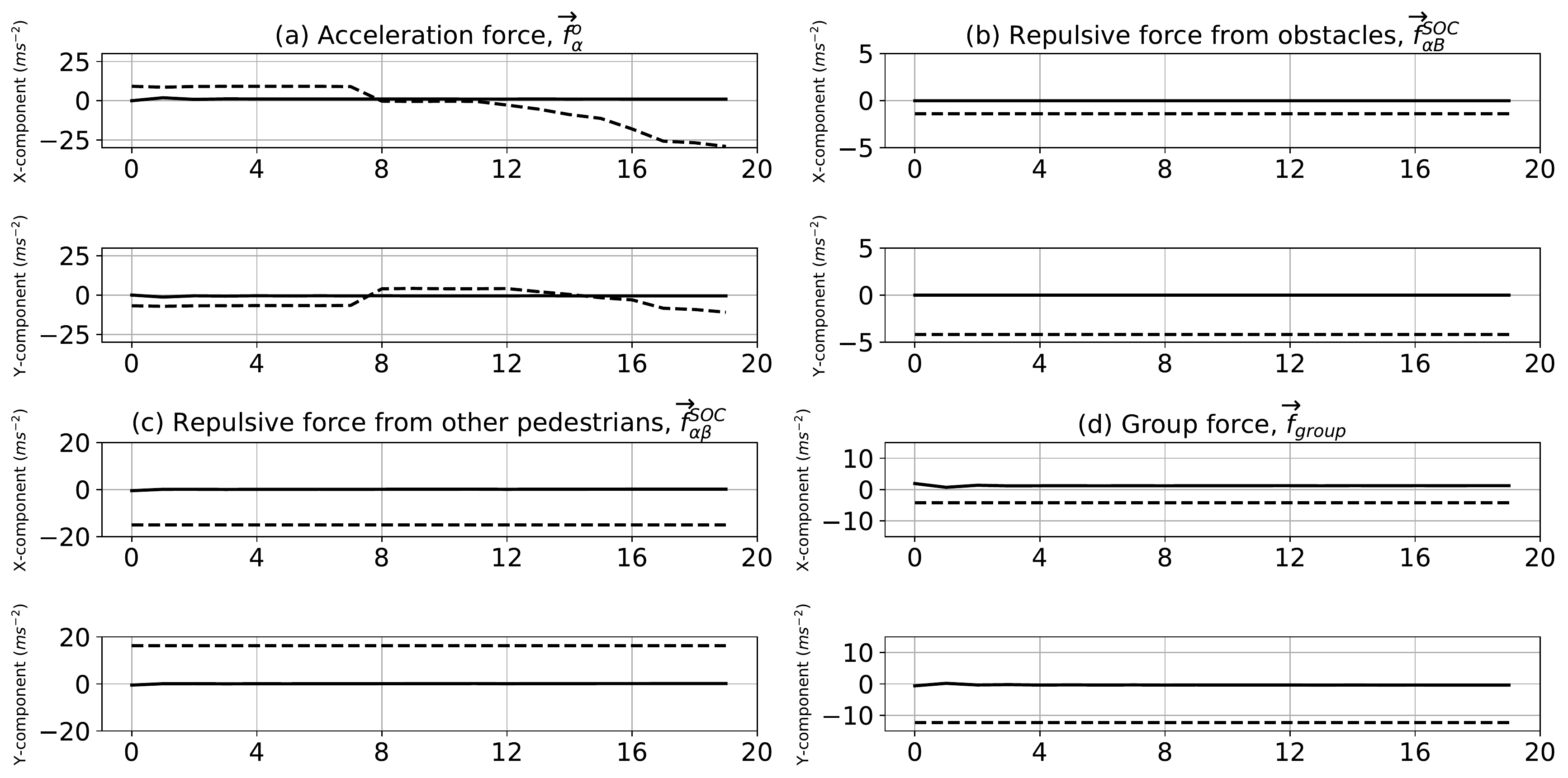}
  \caption{Predicted (dashed line) and simulated (solid line) acceleration force (a), repulsive force from obstacles (b), repulsive forces from other pedestrians (c) and group force (d) acting upon pedestrian 85.}
  \label{fig:individual_forces}
\end{figure}

Figure \ref{fig:individual_forces} shows the individual forces, namely: acceleration force (a), repulsive force from boundaries (b), repulsive from other pedestrians (c) and group force (d). The predicted attraction force (Figure \ref{fig:individual_forces} (a)) changes over time, starting at positive values, falling towards zero at around time-step 12 and continues falling afterwards. This could point towards the pedestrian's need to change its velocity over time to compensate for other environmental facts. That is, it would feel a constant acceleration force in an open environment with no other road users and obstacles. But its not the case here. However, the SFMG simulated acceleration force remains stable throughout. Next, we look at the forces exerted by other actors; static and dynamic. Figure \ref{fig:individual_forces} (b) shows that, the pedestrian feels a constant repulsive force (both predicted and simulated) from boundaries throughout its journey. This corresponds towards to its desire to keep safe distance from boundaries. As the nearest boundary point is quite far from it, so the force values are low. Again, the pedestrian is surrounded by seven other pedestrians, so, it should feel a strong cumulative repulsive force from other pedestrians. This behavior is emulated by the predicted repulsive force from other pedestrians shown in Figure \ref{fig:individual_forces} (c). However, SFMG fails to emulate this properly as it simulates force values nearing zero units. Moreover, 85 is traveling in a group with 84. Therefore, it feels group force influencing it, as we see in Figure \ref{fig:individual_forces} (d). The SFMGNet predicted group forces are much higher (in magnitude) compared to SFMG simulated group forces. The predicted group force better represents 85's desire to stay closer to 84. The nature of group force depends on the combination the visibility force, $f_{vis}$ and attraction force towards group centroid, $f_{att}$. However, we do not consider that level of granularity in this work to avoid further complexity. To conclude, we can establish a high level causality behind a pedestrian's motivation behind its motion behavior by inspecting individual force predictions.   
\subsection{Quantitative comparison}
\label{subsec:quantitative_comparison}
We quantitatively assess our model based on two commonly used \cite{alahi2016social}, \cite{sun2020reciprocal} distance-based parameters: average displacement error (ADE) and mean final displacement error (FDE). ADE refers to the mean of all eucledian distance values between the real and predicted values at each time-steps. FDE is the distance between the last predicted position and last real position. The lower the values of both ADE and FDE parameters, the better the model performance.

\begin{table*}
    \caption{ADE/MDE error (in meters) comparison between SFMGNet and other baseline models.}
    \label{tab:error-comparison}
    \centering
    \begin{tabular}{c c c c c c c c} 
        \hline
        Metric & Dataset & CV & FF & S-LSTM & S-GAN & SFM-NN & SFMGNet \\
        \hline
        \multirow{5}{2.5em}{ADE/FDE} & Hotel & 0.27/0.51 & 1.59/3.12 & 0.15/0.33 & 0.72/1.61 & 0.36/0.82 & \textbf{0.07}/\textbf{0.11} \\ 
        & ETH & 0.58/1.15 & 0.67/1.32 & 0.60/1.31 & 0.81/1.52 & 0.68/1.63 & \textbf{0.10}/\textbf{0.14} \\ 
    & UNIV & 0.46/1.02 & 0.69/1.38 & 0.52/1.25 & 0.60/1.26 & 0.46/1.12 & \textbf{0.35}/\textbf{0.64} \\ 
    & zara01 & 0.34/0.76 & 0.39/0.81 & 0.43/0.93 & 0.34/0.69 & 0.35/0.85 & \textbf{0.11}/\textbf{0.17} \\ 
    & zara02 & 0.31/0.69 & 0.38/0.77 & 0.51/1.09 & 0.42/0.84 & 0.38/0.95 & \textbf{0.13}/\textbf{0.23}
    \\
    \hline
    & Average & 0.39/0.83 & 0.74/1.48 & 0.44/0.98 & 0.58/1.18 & 0.45/1.07 & \textbf{0.15}/\textbf{0.26}
    \\
    \hline
\end{tabular}
\end{table*}

For comparison, we choose constant velocity model (CV), a MLP or feed-forward network based model \cite{scholler2020constant} (FF), S-LSTM model \cite{alahi2016social}, S-GAN model \cite{gupta2018social} and the model described in \cite{antonucci2020generating} (SFM-NN) as baselines. The baseline models (expect SFM-NN) take 3.2 seconds (8 time-steps) of trajectory values as input and predict next 4.8 seconds of trajectory, and report error values. They are also trained on the corresponding real-world datasets. The SFM-NN is trained on synthetic dataset and takes 1 second of data as input to predict 4.8 seconds of trajectory. Our model (SFMGNet) is trained solely on synthetic dataset. It takes 1.2 seconds (as n = 10 chosen) of data as input and predicts the next 4.8 seconds of data. Then, we calculate the error values. The error values are reported in Table \ref{tab:error-comparison}. The bold values represent the lowest or best error values. As we can see our model SFMGNet outperforms every baseline model in every scene dataset. Especially, the lowest ADE and FDE values (i.e. \textbf{0.07} and \textbf{0.11} respectively) are found for \textit{hotel} scene data in ETH dataset. Our model shows worst performance on UNIV dataset: ADE = \textbf{0.35} and FDE = \textbf{0.64}. Still, the model performance is noticeably better than the baseline models. The average ADE and FDE values are: \textbf{0.15} and \textbf{0.26}.

We evaluate SFMGNet's ability to predict feasible and realistic trajectories by calculating the evaluation metric introduced in \cite{8953374}. That is, we calculate the percentage of \textit{near-collisions} among pedestrians for each frame or time-step. A near-collision takes place when the Eucledian distance between the pedestrians is below 0.1 meter \cite{8953374}. We compute the average percentage of human near-collision in each frame of ETH and UCY datasets. Our model (SFMGNet) predicted trajectories does not contain any near-collisions in both datasets. That is, SFMGNet predicts 0.0\% near-collisions for all frames in ETH, ETH hotel, UNIV, zara01 and zara02 scenes. This in turn strongly indicates that, SFMGNet can predict feasible and realistic trajectories.

\subsection{Ablation study}
\label{subsec:ablation_study}
\begin{table*}
    \caption{Evaluation of the ablative models and the proposed model (SFMGNet) based on ADE/MDE (in meters).}
    \label{tab:ablation_study}
    \centering
    \begin{tabular}{c c c c c c} 
        \hline
        Metric & Dataset & Att & ABr & ABrPr & SFMGNet\\
        \hline
        \multirow{5}{3em}{ADE/FDE} & Hotel & 0.63/1.11 & 0.13/0.23 & 0.13/0.21 & \textbf{0.07}/\textbf{0.11} \\ 
        & ETH & 2.92/5.42 & 3.23/5.51 & 2.14/4.06 & \textbf{0.10}/\textbf{0.14} \\ 
    & UNIV & 1.12/2.16 & 0.37/0.67 & 0.37/0.67& \textbf{0.35}/\textbf{0.64} \\ 
    & zara01 & 0.90/1.71 & 0.20/0.34 & 0.15/0.25& \textbf{0.11}/\textbf{0.17} \\ 
    & zara02 & 0.93/1.74 & 0.23/0.41 & 0.15/0.25 & \textbf{0.13}/\textbf{0.23}
    \\
    \hline
    & Average & 1.30/2.43 & 0.83/1.43 & 0.59/1.09 & \textbf{0.15}/\textbf{0.26}
    \\
    \hline
\end{tabular}
\end{table*}

To assess the impact of different modules in our proposed model framework, i.e. acceleration force, repulsive force from static obstacles, repulsive force from other pedestrians, and group force, we test performance of several ablative models. We choose three different model versions, namely: Att, ABr and ABrPr for this study. Where, the model Att considers only acceleration force towards destination, model ABr considers both acceleration force and repulsive force from static obstacles and model ABrPr considers all forces except group force. Finally we compare them with the complete framework, SFMGNet. We test them on ETH and UCY datasets, and report the corresponding ADE/FDE error values in Table \ref{tab:ablation_study}. We can see, the model performance improves with addition of each force estimation module. Model Att performs the worst across all datasets and performance increases significantly when repulsive force from obstacles is considered in model ABr. This trend follows for model ABrPr, which considers repulsive force from other pedestrians. Models Att, ABr and ABrPr perform notably worse on ETH scene compared to the other scenes. However, SFMGNet outperforms these ablative models significantly, indicating towards the strong importance of pedestrian groups.

\paragraph{} Based on the above quantitative and qualitative evaluation, we conclude that, SFMGNet can model realistic, interpretable pedestrian trajectories. Also, it shows better than state-of-the-art performance on two real world benchmark datasets in terms of distance-based metrics.

\section{Conclusion and outlook}
\label{sec:conclusion}
This work proposes a physics-based neural network-based framework namely SFMGNet to predict pedestrian trajectories while considering influences of static obstacles, other pedestrians and pedestrian-groups. SFMGNet combines the social force model with group behavior modeling (SFMG) and multi-layer perceptron (MLP). This combination allows the MLP based model to inherit SFMG model's interpretable nature. This model has the potential to contribute towards planning and understanding of motion behavior of robots and autonomous vehicles. Based on the evaluation results, we conclude that the model predicts realistic trajectories and outperforms state-of-the-art models on ETH and UCY dataset. We can also causally interpret model predictions. However, there remain potential for further improvement. We aim to include other existing road users (e.g. cars, cyclists, etc.) in the model to better emulate heterogeneous mixed-traffic zones. Additionally, we want to explore our model's potential in modeling new types of road users (e.g. cargo-movers, tram buses, etc.), even when relevant data is scarce/absent.

\bibliography{sample-ceur}

\end{document}